%% file: silhouette-neurips.tex
\documentclass{article}

\PassOptionsToPackage{numbers}{natbib}

\usepackage[preprint]{neurips_2022}

\usepackage[utf8]{inputenc} 
\usepackage[T1]{fontenc}    
\usepackage{hyperref}       
\usepackage{url}            
\usepackage{booktabs}       
\usepackage{amsfonts}       
\usepackage{nicefrac}       
\usepackage{microtype}      
\usepackage{xcolor}         
\usepackage{wrapfig}

\title{Silhouette: Toward Performance-Conscious \\ and Transferable CPU Embeddings}

\author{%
  Tarikul Islam Papon \\
  Boston University \\
  \texttt{papon@bu.edu} \\
  \And
  Abdul Wasay \\
  Intel Labs \\
  \texttt{abdul.wasay@intel.com} \\
}

\input{./files/custom}

\begin{document}

\maketitle

\begin{abstract}

Learned embeddings are widely used to obtain concise data representation and enable transfer learning between different data sets and tasks. In this paper, we present Silhouette, our approach that leverages publicly-available performance data sets to learn CPU embeddings. We show how these embeddings enable transfer learning between data sets of different types and sizes. Each of these scenarios leads to an improvement in accuracy for the target data set.  
  
\end{abstract}


\input{files/1_intro}
\input{files/2_related}

\input{files/1_main}
\input{files/3_exp}

\section{Conclusion} We present Silhouette, a performance-conscious learned embedding for CPUs. We show how we can use Silhouette for transfer learning across data sets of different sizes and types. In these scenarios, Silhouette consistently improves accuracy. 

\boldPara{Limitations and Future Work.} The current study only considers the SPEC CPU 2017 data set, a fully-connected neural network model, and a regression prediction task. There are opportunities to consider other data sets, including data sets generated from micro-architecture simulators. Similarly, we can consider other configurations of models (sequence models, decision trees, etc.) and tasks (classification, clustering, design generation, etc.).



\bibliographystyle{plainnat}
\bibliography{bibliography.bib}

\appendix

\input{files/appendix}

\end{document}

%% file: files/custom.tex
\usepackage{color}
\usepackage{bm}
\usepackage{xspace}
\usepackage{subcaption}
\usepackage[disable]{todonotes}
\usepackage{tabularx}

\newcommand{\boldPara}[1]{\vspace{0.05in}\noindent \textbf{#1}}

\newcommand\rev[1]{\textcolor{red}{#1}} 

\renewcommand\rev[1]{#1}
\newcommand{\boldC}[0]{\bm{C}}
\newcommand{\boldD}[0]{\bm{D}}

\newcommand{\boldV}[0]{\bm{V}}

\newcommand{\boldX}[0]{\bm{X}}
\newcommand{\boldY}[0]{\bm{Y}}

%% file: files/1_intro.tex
\section{Introduction}

\boldPara{Widespread Learned Embeddings.} 
Learned embeddings transform high-dimensional data sets into a low-dimensional space while preserving semantic and relational information. For instance, Word2Vec is a natural language embedding that converts words into vectors. The distance between vectors represents how close or far-off the corresponding words are in their meanings. The machine learning community has designed and used such embeddings for diverse data types, including network graphs, images, and chemical molecules \citep{chen:2018:networktutorial}. 

\boldPara{Embeddings Enable Transfer Learning.}
Embeddings provide a concise way of capturing high-level relationships and patterns in data that generalize to other scenarios. In this way, embeddings enable transfer learning between different data sets and tasks. For instance, various approaches often use an embedding trained for image classification on the ImageNet data set to improve accuracy on related tasks for which large data sets might not exist (such as detecting agriculture pests) \cite{huh:2016:imagenet_transfer}. 

\boldPara{Embeddings and CPU Performance.} Prior research has successfully designed embeddings for various machine learning tasks, including hardware-related tasks such as developing and verifying Application-Specific Integrated Circuits (ASICs); however, performance-conscious and transferable learned embeddings do not exist for general-purpose CPUs. This makes it hard to transfer knowledge between different data sets and learning tasks within the CPU-performance regime, such as performance prediction, CPU selection, and CPU ranking. This is crucially problematic because only a limited number of data sets contain large-scale CPU performance profiles, i.e., standardized data with performance across several generations of CPUs. 

\boldPara{Silhouette.}
We present Silhouette, performance-conscious and transferable CPU embedding that converts a CPU specification into a low-dimensional and continuous vector space while capturing the CPU's performance profile. Silhouette is trained on a regression task i.e., to predict normalized performance on the SPEC CPU 2017 performance data set. We show how Silhouette enables transfer learning between data sets of different types and sizes. Crucially, we can use Silhouette to improve accuracy for data sets and tasks with less number of training samples. 

\boldPara{Contributions.} We make the following contributions:

\begin{itemize}
    \item We design Silhouette, a novel performance-conscious embedding for CPUs that converts CPU specifications into a continuous vector space while capturing its performance properties. 
   
   \item We show how we can integrate various sources of publicly-available data sets (Intel's Ark database and SPEC's CPU 2017 benchmark results) to create a rich training data set to train Silhouette. Further, we plan to make this integrated data set available to the research community.

   \item \rev{We show how we can use Silhouette to do transfer learning to improve accuracy across different data sets. We show how Silhouette can provide up to $6\times$ improvement in accuracy when used to predict SPEC 2017 performance numbers for CPU architectures having a smaller number of training samples. We also show how Silhouette trained on Intel processors can help with improving prediction accuracy on non-Intel processors.}
\end{itemize}

%% file: files/2_related.tex
\section{Related Work}
\boldPara{Hardware Embedding}
Mathematical and learned embeddings are proposed to enhance various tasks such as neural network compilation and chip design for GPUs and ASICs \citep{ahn:2022:glimpse,lee:2021:help,vasudevan:2021:design2vec}: 
Glimpse applies principal component analysis to a GPU specification to create an embedding that enables transfer learning between different GPUs for neural compilation \citep{ahn:2022:glimpse}. 
HELP proposes to record the latency of a given GPU on a set of benchmarks and use these latency numbers as the GPUs embedding \citep{lee:2021:help}. 
Finally, Design2Vec introduces a learned embedding to transform the register transfer language (RTL) description of TPUs into a continuous space. This continuous space enables better design exploration \citep{vasudevan:2021:design2vec}. 
\rev{Silhouette develops performance-based embeddings for the complex design space of general-purpose CPUs. These embeddings capture the performance characteristics of various CPU designs, and we show how we can use them for transfer learning across different data sets and tasks.}

\boldPara{Performance Prediction Models.}
There are various approaches to predict performance on CPUs and they utilize mechanistic models \citep{chen:2011:hybrid,van:2015:micro} as well as empirical machine learning models \citep{singh:2007:predicting,lopez:2018:predict,wang:2019:predictcpuspec}. 
Closely related to Silhouette are recent efforts to use machine learning models to predict performance on SPEC CPU 2006 performance data set \citep{lopez:2018:predict,wang:2019:predictcpuspec}: Approaches train both linear and neural network regression models and apply them to rank CPU designs for consumers. 
\rev{Silhouette extends these approaches by introducing a performance-based CPU embedding. We train this embedding using SPEC CPU 2017 data set on prediction task. We show how leveraging these embeddings can improve prediction accuracy for CPUs with a lower number of training samples.}

%% file: files/1_main.tex
\section{Silhouette: Model Design and Training}

Silhouette operates by taking the specification of a CPU design $\boldC$, which contains features that describe different aspects of the CPU (e.g., clock speed, cache sizes, etc.\footnote{Table \ref{tab:data} provides a complete list of features we use in our model.}) and outputs a k-dimensional continuous vector $\boldV \in \mathbb{R}$. $\boldV$ is an embedding for the CPU design $\boldC$.

$$
\boldV = S(\boldC)
$$

\textbf{Embedding function.} The embedding function takes the form of a fully-connected neural network, with an input of size $|\boldC|$ and output of size $|\boldV|$.

\textbf{Training task.}
This embedding function $S$ is trained on a regression task. To do so, we attach a predictor sub-network $P$ to the embedding function to create a neural network $N(\boldC)$ = $P(S(\boldC))$. We train $N$ using the training data set
$\boldD = \{\boldX, \boldY\}$ such that $\boldX = \{ \boldC_0 ... \boldC_{n-1}\}$ and $\boldY$ is a corresponding vector of targets. 

\textbf{Training data.} We train Silhouette on a data set derived by integrating two publicly-available sources of data: SPEC CPU 2017 performance data and Intel Ark. SPEC CPU 2017 is the leading industry benchmark suite to analyze and report CPU performance. It consists of 43 benchmarks that map to diverse workloads ranging from physics to artificial intelligence to molecular biology. SPEC CPU 2017 hosts a data repository with the performance (reported as speed and latency) of various CPU configurations on these benchmarks. The SPEC data set has limited detail about CPUs, and we integrate this data set with more CPU features from the Intel Ark data set. Overall this results in around $50K$ training samples containing unique $286$ Intel and $95$ non-Intel CPUs with different configurations. We provide a detailed analysis of this data set and prepossessing steps in Appendix \ref{app:dat}.

\noindent
Overall, this results in $50K$ training samples. The input to our model $\boldC_i$ is a vector with $19$ features of different types -- numerical, categorical, and binary. We train our model to predict the normalized performance on the SPEC CPU 2017 benchmark. 

%% file: files/3_exp.tex
\section{Experimental Evaluation}
\label{sec:exp}

\begin{figure}[t]
        \captionsetup{width=.4\linewidth}
        \begin{minipage}{0.49\textwidth}
		\includegraphics
		[width=\textwidth]
		{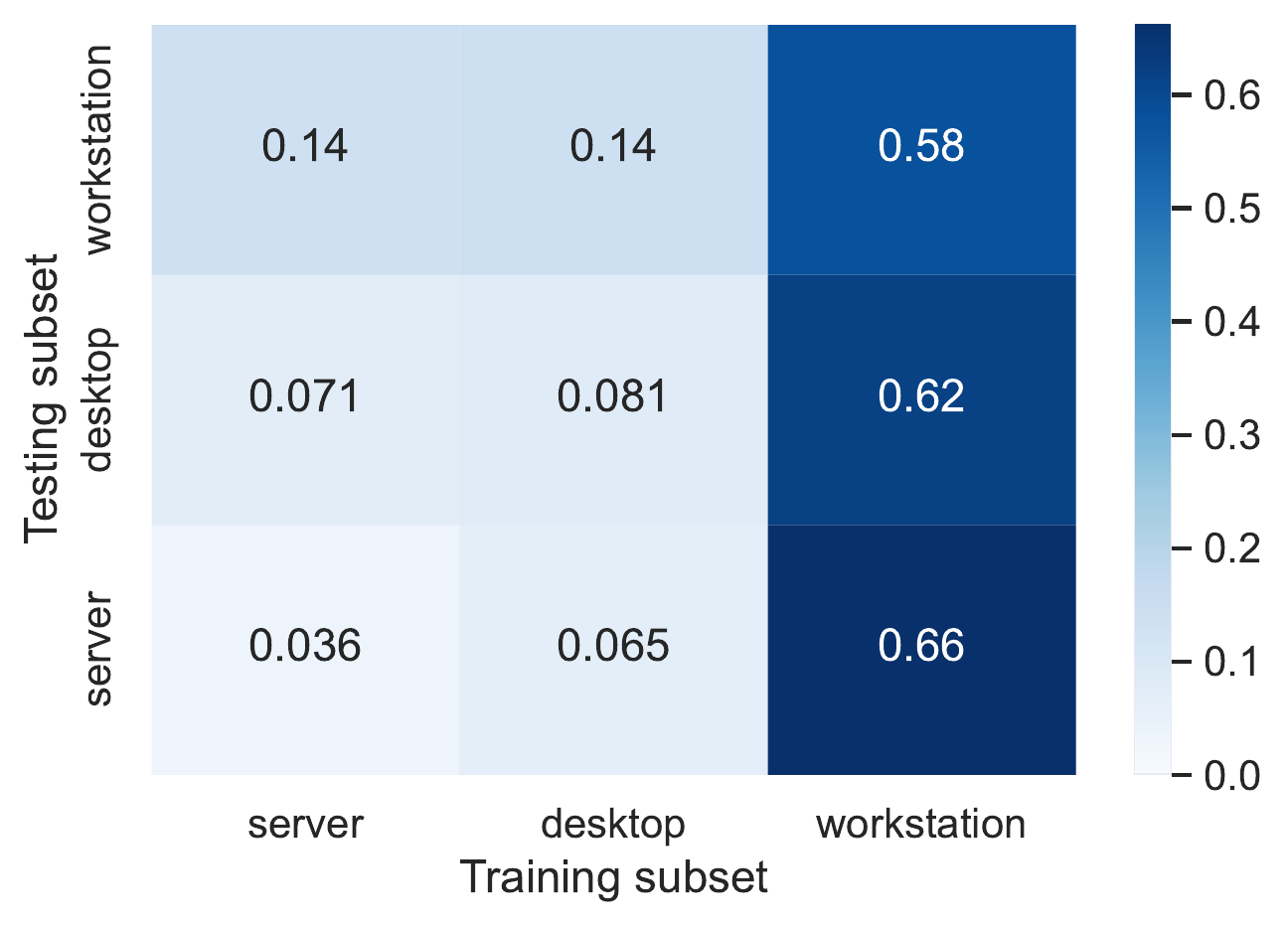}
		\caption{Mean absolute error across every pair of product type. \label{fig:a2a_product_type}}
	\end{minipage}
        \begin{minipage}{0.49\textwidth}
		\includegraphics
		[width=\textwidth]
		{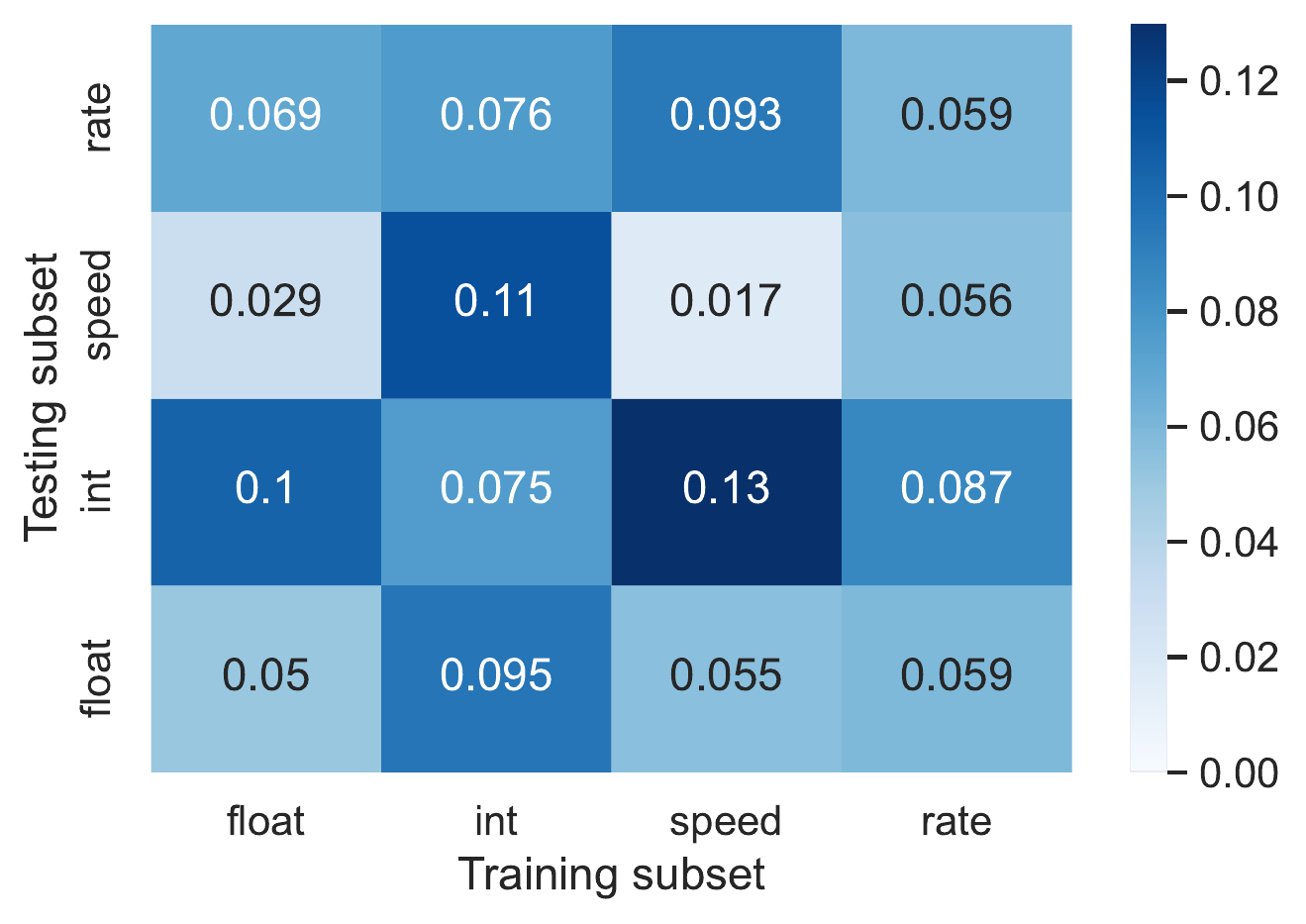}
            
		\caption{Mean absolute error across every pair of benchmark type.\label{fig:a2a_benchmark_type}}
	\end{minipage}
	
 \end{figure}
 \begin{figure}[t]
  \captionsetup{width=.4\linewidth}
	\begin{minipage}{0.49\textwidth}
		\includegraphics
		[width=\textwidth]
		{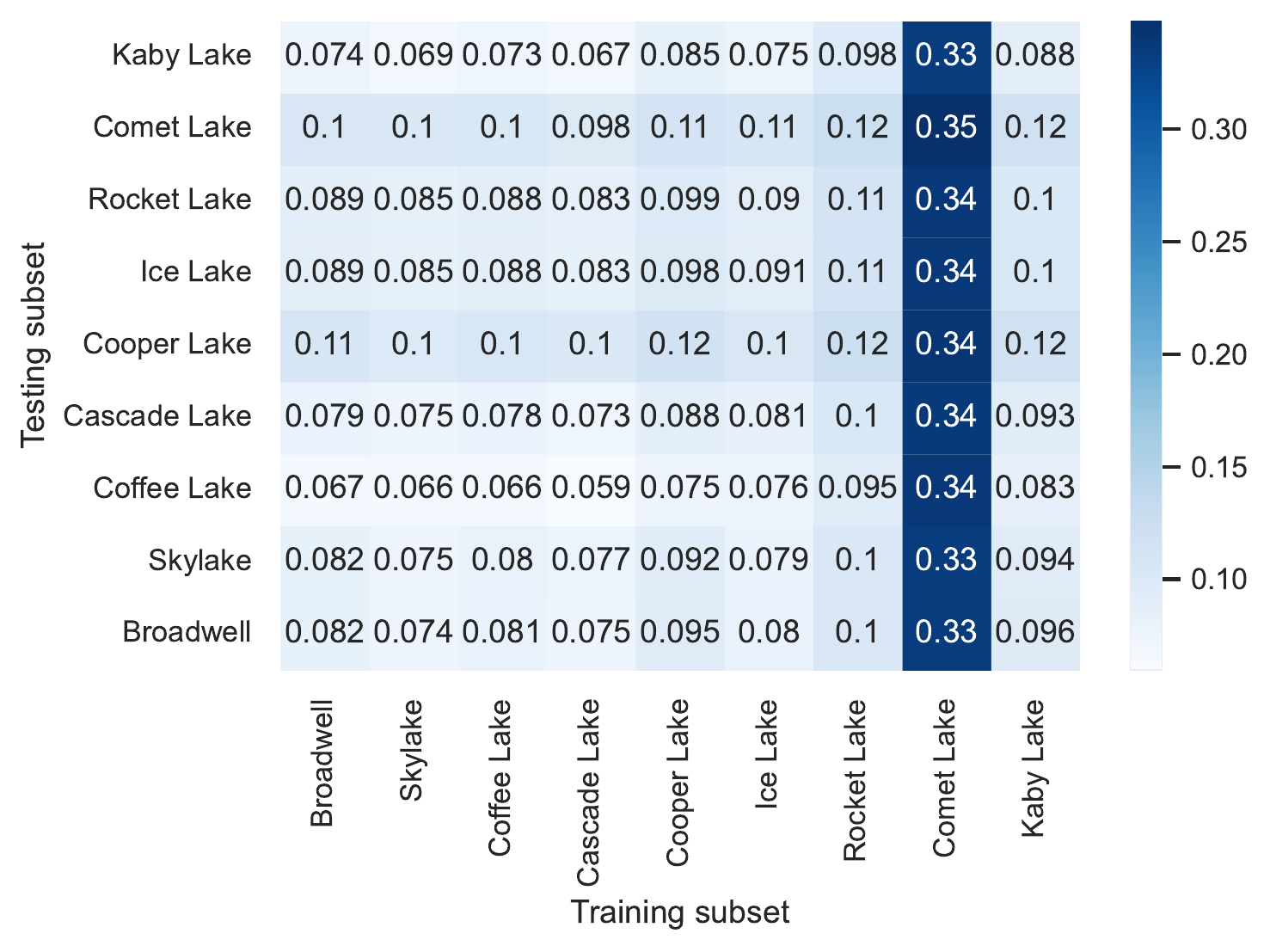}
		\caption{Mean absolute error across every pair of architecture type. \label{fig:a2a_architecture_type}}
	\end{minipage}
	\begin{minipage}{0.465\textwidth}
		\includegraphics
		[width=\textwidth]
		{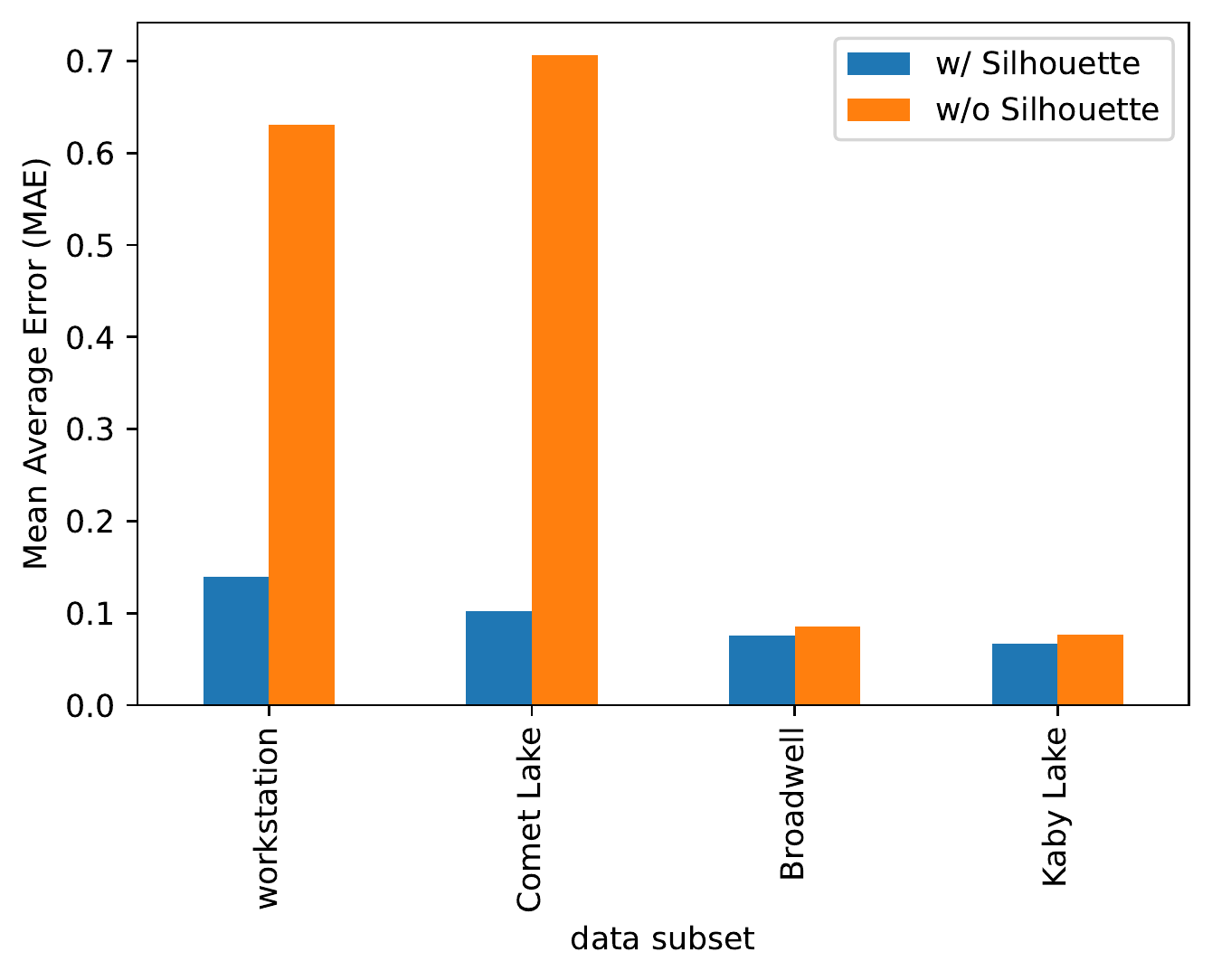}
		\caption{Silhouette improves accuracy for data sets with low number of samples. \label{fig:transfer}}
	\end{minipage}
 \vspace{-0.3in}
\end{figure}

Taking inspiration from past work on language embeddings, we evaluate Silhouette by showing how it can enable transfer learning between different data sub-sets and tasks. 

\boldPara{Experimental setup.} We evaluate various configuration of the hyperparameters to select an optimal set: we train all models using RMSprop with a mini-batch size of 64, a learning rate of 0.001, and 0.9 momentum. We randomly shuffle the training data before every training epoch and all weights are initialized by sampling from a normal distribution. All models are trained till they converge. All experiments are repeated ten times and we report the average Mean Absolute Error (MAE). We use a fully-connected model with three hidden layers each of size 100 and the Sigmoid activation function. 


\boldPara{Transfer between different subsets of data.} First, we show how Silhouette trained on one data set performs on other data sets. 
We look at three divisions of the SPEC CPU 2017 data set. We create these divisions based on three attributes: (i) product type (server, desktop, or workstation), (ii) type of benchmark (Rate, Speed, Int, or Float), and (iii) architecture type (Broadwell, Skylake, etc.). Each of these divisions contains the entire data set divided into subsets; every subset contains data samples corresponding to one value of the selected attributes. For instance, based on the `product type' attribute, we divide the data into three subsets, each corresponding to one of the three product types: server, desktop, or workstation. Table \ref{tab:product_type}, Table \ref{tab:benchmark_type}, and Table \ref{tab:intel_sku} in the Appendix shows how many subsets correspond to every division and the number of training samples in every subset. 
Table \ref{tab:data} list the details of data used during training and testing.

We train Silhouette separately on data corresponding to one of the attribute values and test its performance on data corresponding to the rest of the attributes. We report the Mean Absolute Error (MAE) for every attribute pair in Figure \ref{fig:a2a_product_type}, \ref{fig:a2a_benchmark_type}, and \ref{fig:a2a_architecture_type}. The x-axis indicates the data subset on which we train, and the y-axis indicates the data subset on which we test.
\rev{Overall, we observe that across all these data subsets, Silhouette shows a high degree of transferability and achieves low MAE in the range of $0.02$ to $0.7$. For data subsets with high number of samples such as Server, Skylake, and Cascade Lake, we observe high accuracy (i.e., low MAE) whereas for subsets with less number of samples, we observe lower MAE.}.

\boldPara{Transfer from large to small data sets.} In the next experiment, we show how Silhouette can be used to improve accuracy for data sets having a very low number of training samples. This property of Silhouette is particularly important since there is an assymtery of data set between different processors. We pick four subsets $s$ of the SPEC CPU 2017 data: workstation (product type) and three Intel architecture types: Kaby Lake, Comet Lake, and Broadwell. These subsets have an order of magnitude less training samples compared to other subsets. We train and evaluate two prediction models: (i) w/o Silhouette and (ii) w/ Silhouette. In the first case, we train and evaluate a model using just the data samples corresponding to the subset $s$. In the second case, we replace the input to the model with Silhouette trained on the entire data set except $s$. 

We report the results in Figure \ref{fig:transfer}. We observe that for these data sets, Silhouette leads to a considerable improvement in accuracy (lower MAE). The improvement is higher for Workstation and Cascade Lake data subsets as they have significantly less number of training samples when compared with all other subsets in the data.  

  
\textbf{Transfer from Intel to non-Intel CPUs.} In this experiment, we evaluate whether Silhouette trained on Intel processors can be used to predict the performance of Non-Intel processor. 
\begingroup
\setlength{\columnsep}{0.12in}%
\begin{wrapfigure}{r}{0.5\textwidth}
	\centering
	\vspace{-0.14in}
    \includegraphics[width=1\linewidth]{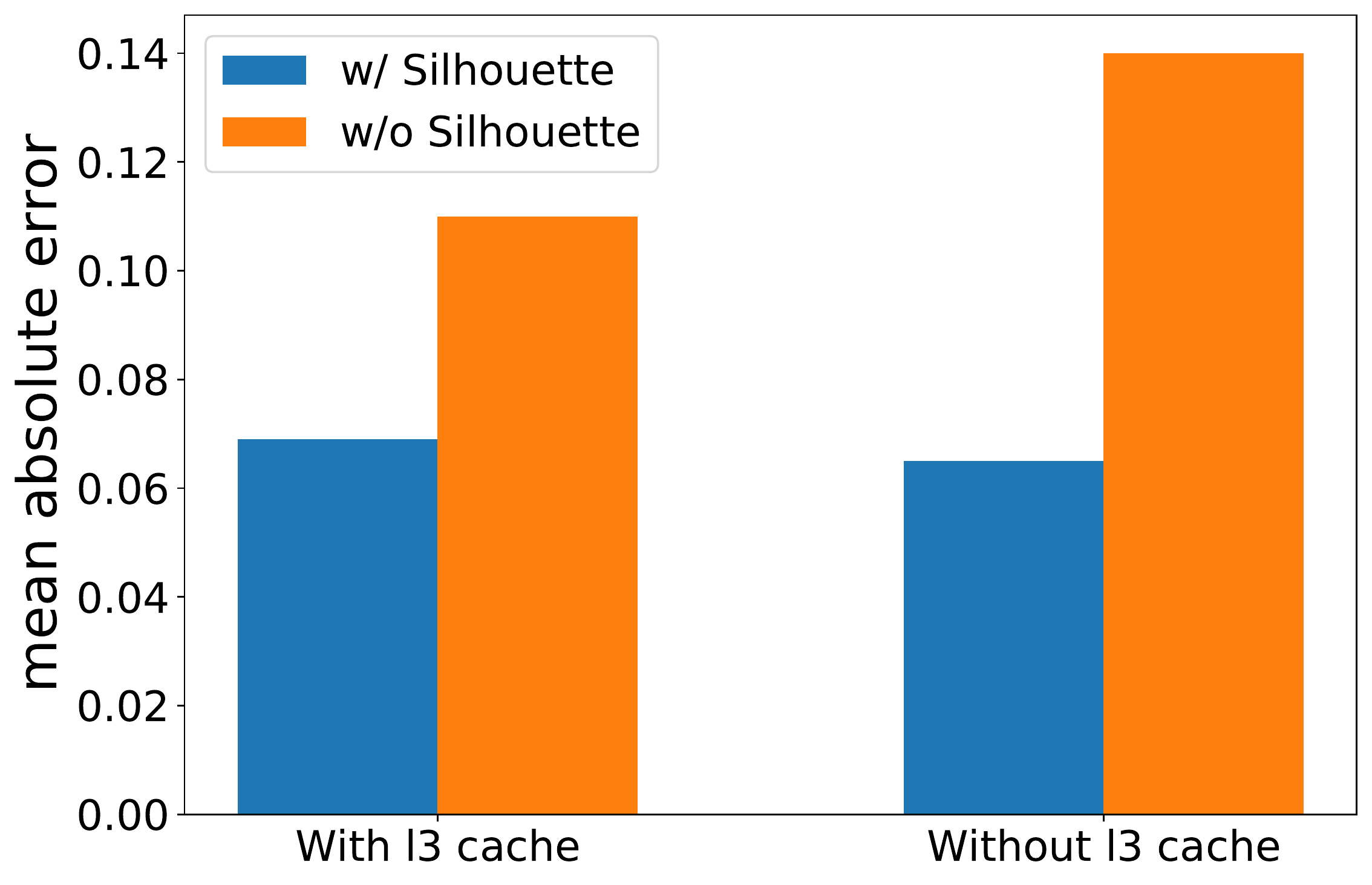}
	\vspace{-0.2in}
	\caption{\small Silhouette trained on Intel data set can be used to improve prediction accuracy on non-Intel processors.}
	\vspace{-0.1in}
  \label{fig:nonintel}
\end{wrapfigure}
First, we remove the Intel-specific features (e.g., turbo-boost technology, hyper-threading, etc.) from the training data which contains only Intel processors. Then, we train Silhouette on this data. To prepare the testing data, we select the 15 most frequent non-Intel processors in the SPEC data for different configurations ($\sim$1000 testing data) and crawl the processors' detailed specifications. We apply the same data pre-processing steps on the testing data.
The result of this experiment is presented in Figure~\ref{fig:nonintel}. Note that, non-Intel processors and Intel processors have very different l3 cache designs, hence, in general, the non-Intel processors have a significantly larger cache size. We experiment both with and without l3 cache size for completeness. The results shows that (i) while Silhouette is trained on the Intel data set only, it can predict the performance of non-Intel processors with low mean absolute error, and (ii) since the l3 cache values are quite different by nature in Intel and non-Intel processors, this feature can be excluded for such a use case.

%% file: files/appendix.tex
\section{Training Data set}
\label{app:dat}
\textbf{Intel Processor Specifications.} We collect the detailed Intel processor specification dataset from \url{http://ark.intel.com} where all the specifications of Intel’s processors are publicly available. 
We wrote a crawler that collected 1500+ processors' specifications which include microarchitecture, product type, launch year, number of cores, frequency etc. 
Note that not all processors have the same set of features. 
In other words, different processors have different features depending on their nature and we found more than 120 different features across all the processors.
However, some crucial features are common across all the processors. 
The common features that were eventually selected for the training are: microarchitecture, type, l3 cache size, instruction set architecture, memory type, channel count, ecc supported, base frequency, turbo frequency, turbo boost technology, total cores, total threads, hyperthreading, tdp, and release year.

\begin{figure}[t]
  \centering
  \begin{subfigure}{.46\linewidth}
    \centering
    \includegraphics[width=\linewidth]{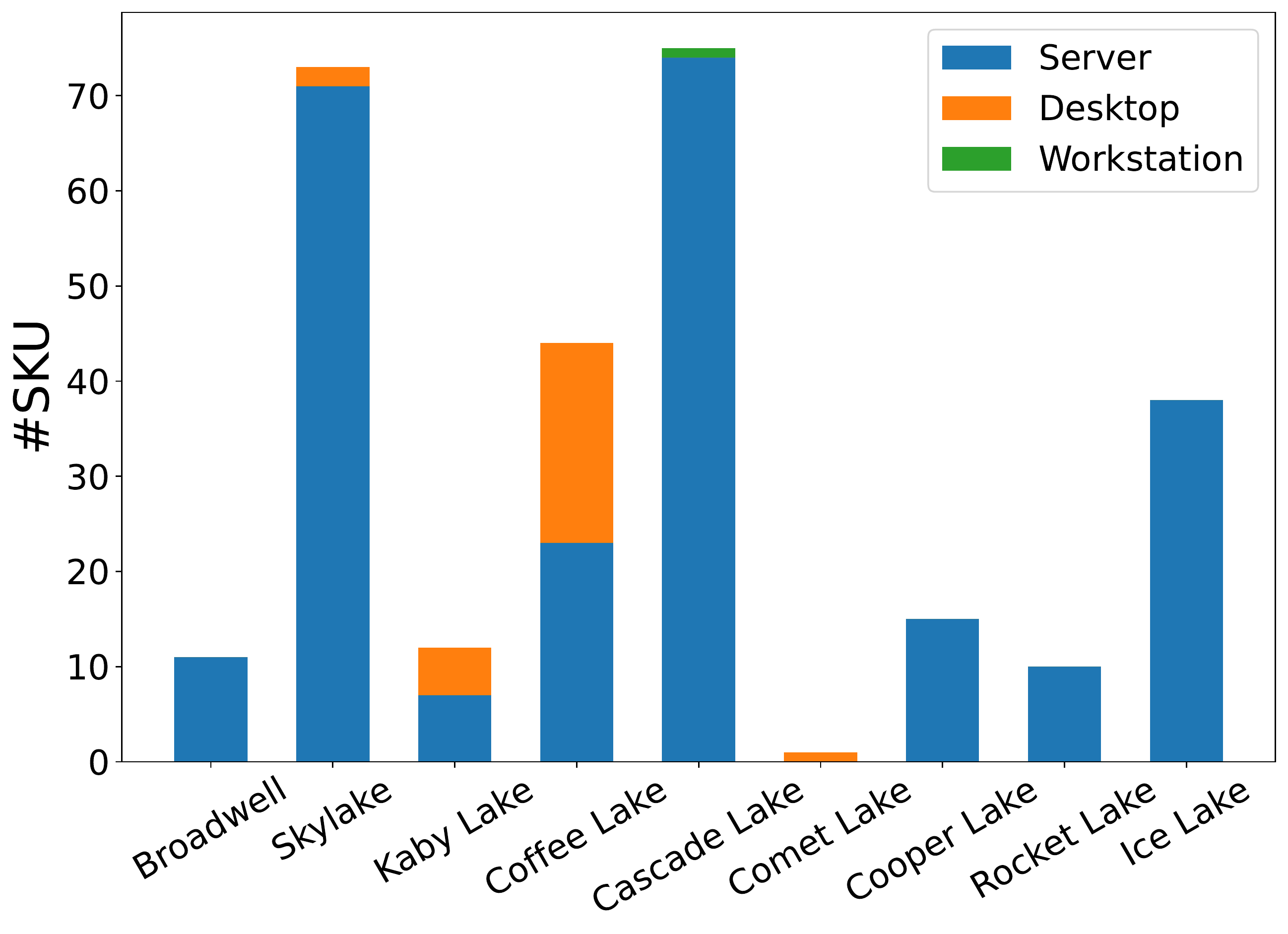}
    \vspace{-0.18in}
    \caption{}
    \label{fig:microarchi}
  \end{subfigure}%
  \enskip
  \begin{subfigure}{.52\linewidth}
    \centering
    \includegraphics[width=\linewidth]{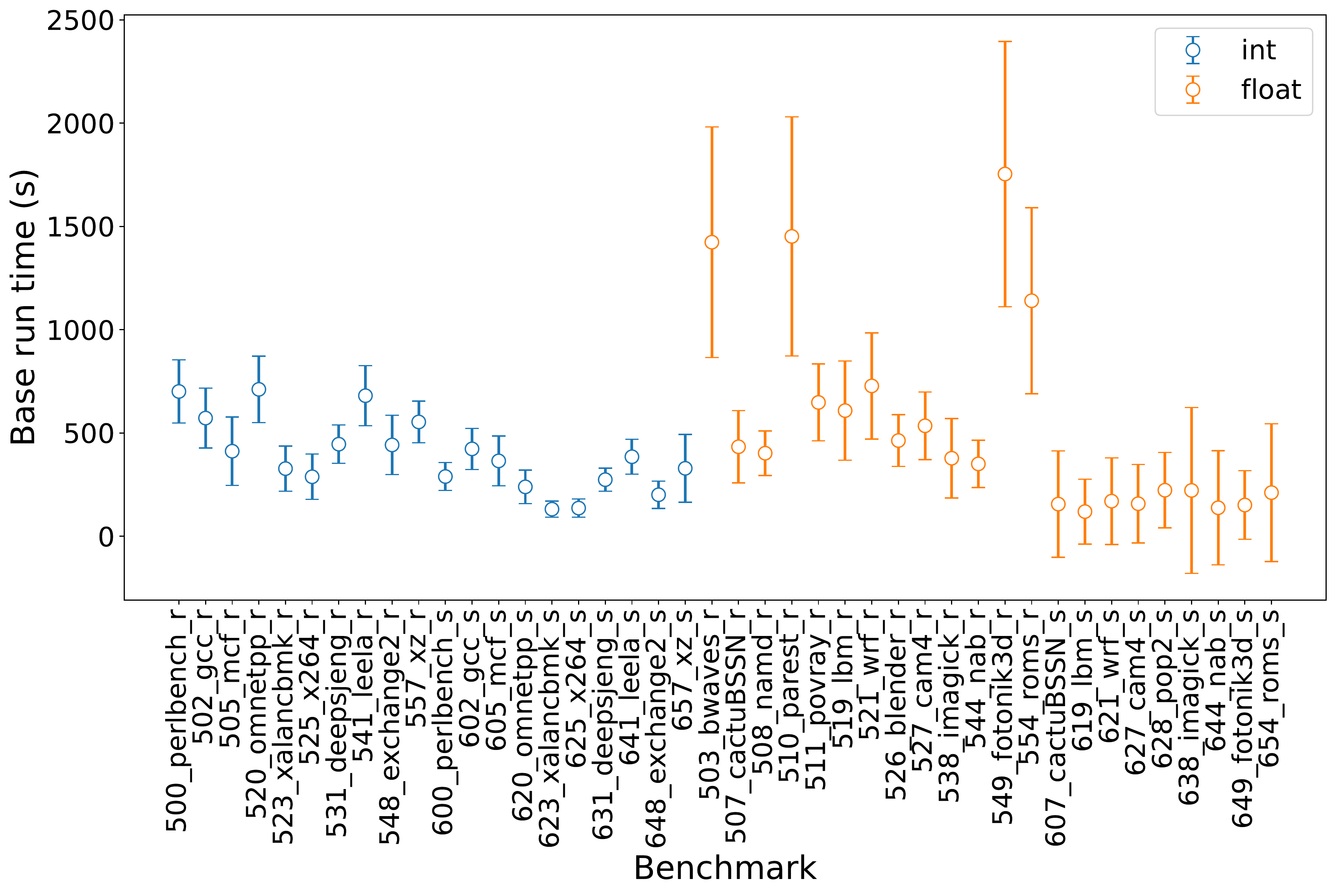}
    \vspace{-0.18in}
    \caption{}
    \label{fig:float_int}
  \end{subfigure}
  \vspace{-0.1in}
  \caption{(a) Distribution of Intel SKUs based on their microarchitecture and product type. (b) Floating point benchmarks are more diverse than integer point benchmarks.}
  \vspace{-0.05in}
\end{figure}

\textbf{SPEC CPU 2017.} SPEC CPU 2017 focuses on compute-intensive performance, which means these benchmarks emphasize the performance of processor, memory and compilers. 
SPEC CPU includes 4 suits that focus on different types of compute-intensive performance consisting of both integer and floating point microbenchmarks.
In total, SPEC has 43 microbenchmarks (23 belonging to floating points and 20 belonging to integer benchmarks).
The benchmark reports the base run time and peak run time for a processor for a certain configuration.
A CPU configuration is an SKU running at a certain frequency with a certain memory size and with a certain number of enabled threads.
A configuration (SKU, enabled core, thread count, memory size) determines a workload’s performance. 
The benchmark also reports two other metrics: SPECspeed and SPECrate which we did not use for our evaluation.

After running SPEC CPU 2017, we found the performance numbers for 286 Intel processors (all released after 2015) and 95 non-Intel processors.
Figure~\ref{fig:microarchi} presents the distribution of the processors
\begingroup
\setlength{\columnsep}{0.09in}%
\begin{wrapfigure}{r}{0.44\textwidth}
	\centering
	\vspace{-0.14in}
    \includegraphics[width=1\linewidth]{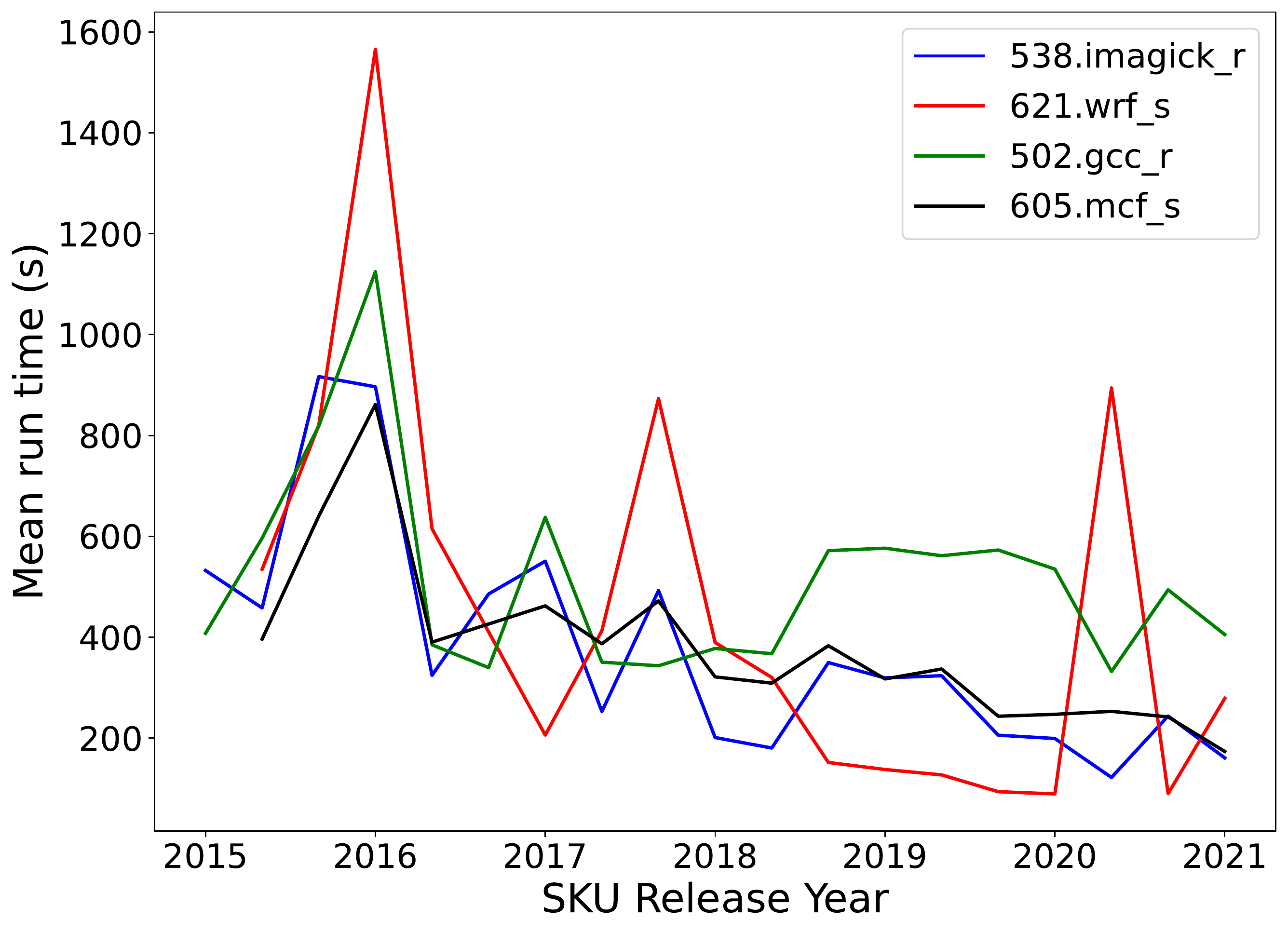}
	\vspace{-0.25in}
	\caption{\small Lower runtime for recent SKUs}
	\vspace{-0.15in}
  \label{fig:trend}
\end{wrapfigure}
based on their microarchitecture and product type.
The figure shows that most of the SKUs are of `Server' type and a majority of the SKUs belong to Skylake and Cascade Lake microarchitecture.
As for the benchmarks, we analyzed their mean and standard deviation across all processors and configurations. 
Figure~\ref{fig:float_int} shows this analysis which reveals that in general, the floating point benchmarks are more diverse than the integer benchmarks.
The performance trend of four representative benchmarks from 4 suits based on SKU release year is presented in Figure~\ref{fig:trend}.
The figure shows that in general the average run time is getting lower with newer SKUs and this trend remains true across all benchmarks.


\textbf{Data Preprocessing.} The SPEC datasets contains a
configuration (SKU, enabled core, thread count, memory size), the workload name, and performance (base runtime in seconds, which is the value to predict). 
We then fetch the detailed CPU specifications from the crawled Intel processor specifications based on the corresponding SKU processor number.
Numerical inputs (e.g. l3 cache size, frequency, etc.) are kept numerical while categorical inputs (e.g. product type, microarchitecture, etc.) are integer encoded. 
We eliminate the duplicate entries first, then, we calculate the average runtime of the workloads with the same configuration. The runtime is then normalized between 0 and 1.



\begin{table}[h!]
    \centering
    \caption{Number of samples corresponding to every product type in the SPEC CPU 2017 data set.}
    \vspace{0.1in}
    \begin{tabular}{lr}
        \toprule
          Product type &  No. of samples \\
        \midrule
             server &                       53206 \\
            desktop &                        1253 \\
        workstation &                          43 \\
        \bottomrule
    \end{tabular}
    \label{tab:product_type}
\end{table}

\begin{table}[h!]
    \centering
    \caption{Number of samples corresponding to every benchmark type in the SPEC CPU 2017 data set.}
    \vspace{0.1in}
    \begin{tabular}{lr}
        \toprule
        Benchmark type &  No. of samples \\
        \midrule
            float &                       28612 \\
              int &                       25890 \\
            speed &                       25320 \\
             rate &                       29182 \\
        \bottomrule
    \end{tabular}
    \label{tab:benchmark_type}
\end{table}

\begin{table}[h!]
    \centering
    \caption{Number of samples corresponding to every Intel SKU in the SPEC CPU 2017 data set.}
    \vspace{0.1in}
    \begin{tabular}{lr}
        \toprule
           Intel SKU &  No. of samples \\
        \midrule
           Broadwell &                         465 \\
             Skylake &                       17308 \\
         Coffee Lake &                        4105 \\
        Cascade Lake &                       19871 \\
         Cooper Lake &                        2249 \\
            Ice Lake &                        8644 \\
         Rocket Lake &                        1401 \\
          Comet Lake &                          43 \\
           Kaby Lake &                         416 \\
        \bottomrule
    \end{tabular}
    \label{tab:intel_sku}
\end{table}

\begin{table}[h]
{
\begin{center}
\caption{We use $19$ input features in our model that capture various aspects of the CPU configuration.}
\vspace{0.1in}
\begin{tabular}{ccc}
\toprule
Input & Details & Type  \\
\midrule
Workload      & Name of the workload            & Categorical           \\
Microarchitecture        & Intel code names representing its microarchitecture            & Categorical           \\
Type        & Product Type (Server, Desktop, Workstation)           & Categorical           \\
L3 Cache Size     & Size of last level cache            & Numerical         \\
Instruction Set Extensions     & SSE or AVX or Both            & Categorical         \\
Memory Type     & DDR3 or DDR4 or Both            & Categorical         \\
Memory Channel Count     & Number of memory channels            & Numerical         \\
ECC Support     & Whether ECC memory is supported            & Binary         \\
Base Frequency     & Base Frequency            & Numerical         \\
Turbo Frequency     & Turbo (Maximum) Frequency            & Numerical         \\
Turbo Boost Technology     & Whether Turbo Boost Technology is supported            & Binary         \\
Total Cores     & Number of cores            & Numerical         \\
Total Threads     & Number of threads            & Numerical         \\
Hyper-Threading     & Whether hyper-threading is supported            & Binary         \\
TDP     & Thermal design power            & Numerical         \\
Year     & Release year            & Numerical         \\
Enabled Cores     & Cores enabled during benchmarking            & Numerical         \\
Thread Count     & Thread count during benchmarking            & Numerical         \\
Memory Size     & Host machine's memory size during benchmarking            & Numerical         \\

\bottomrule
\end{tabular}
\label{tab:data}
\end{center}
}
\vspace{-0.2in}
\end{table}


%% file: silhouette-neurips.bbl
\begin{thebibliography}{10}
\providecommand{\natexlab}[1]{#1}
\providecommand{\url}[1]{\texttt{#1}}
\expandafter\ifx\csname urlstyle\endcsname\relax
  \providecommand{\doi}[1]{doi: #1}\else
  \providecommand{\doi}{doi: \begingroup \urlstyle{rm}\Url}\fi

\bibitem[Ahn et~al.(2022)Ahn, Kinzer, and Esmaeilzadeh]{ahn:2022:glimpse}
Byung~Hoon Ahn, Sean Kinzer, and Hadi Esmaeilzadeh.
\newblock Glimpse: mathematical embedding of hardware specification for neural
  compilation.
\newblock In \emph{Proceedings of the 59th ACM/IEEE Design Automation
  Conference}, pages 1165--1170, 2022.

\bibitem[Chen et~al.(2018)Chen, Perozzi, Al-Rfou, and
  Skiena]{chen:2018:networktutorial}
Haochen Chen, Bryan Perozzi, Rami Al-Rfou, and Steven Skiena.
\newblock A tutorial on network embeddings.
\newblock \emph{arXiv preprint arXiv:1808.02590}, 2018.

\bibitem[Chen and Aamodt(2011)]{chen:2011:hybrid}
Xi~E Chen and Tor~M Aamodt.
\newblock Hybrid analytical modeling of pending cache hits, data prefetching,
  and mshrs.
\newblock \emph{ACM Transactions on Architecture and Code Optimization (TACO)},
  8\penalty0 (3):\penalty0 1--28, 2011.

\bibitem[Huh et~al.(2016)Huh, Agrawal, and Efros]{huh:2016:imagenet_transfer}
Minyoung Huh, Pulkit Agrawal, and Alexei~A Efros.
\newblock What makes imagenet good for transfer learning?
\newblock \emph{arXiv preprint arXiv:1608.08614}, 2016.

\bibitem[Lee et~al.(2021)Lee, Lee, Chong, and Hwang]{lee:2021:help}
Hayeon Lee, Sewoong Lee, Song Chong, and Sung~Ju Hwang.
\newblock Hardware-adaptive efficient latency prediction for nas via
  meta-learning.
\newblock \emph{Advances in Neural Information Processing Systems},
  34:\penalty0 27016--27028, 2021.

\bibitem[Lopez et~al.(2018)Lopez, Guynn, and Lu]{lopez:2018:predict}
Leonardo Lopez, Michael Guynn, and Meiliu Lu.
\newblock Predicting computer performance based on hardware configuration using
  multiple neural networks.
\newblock In \emph{2018 17th IEEE International Conference on Machine Learning
  and Applications (ICMLA)}, pages 824--827. IEEE, 2018.

\bibitem[Singh et~al.(2007)Singh, Ipek, McKee, de~Supinski, Schulz, and
  Caruana]{singh:2007:predicting}
Karan Singh, Engin Ipek, Sally~A McKee, Bronis~R de~Supinski, Martin Schulz,
  and Rich Caruana.
\newblock Predicting parallel application performance via machine learning
  approaches.
\newblock \emph{Concurrency and Computation: Practice and Experience},
  19\penalty0 (17):\penalty0 2219--2235, 2007.

\bibitem[Van~den Steen et~al.(2015)Van~den Steen, De~Pestel, Mechri, Eyerman,
  Carlson, Black-Schaffer, Hagersten, and Eeckhout]{van:2015:micro}
Sam Van~den Steen, Sander De~Pestel, Moncef Mechri, Stijn Eyerman, Trevor
  Carlson, David Black-Schaffer, Erik Hagersten, and Lieven Eeckhout.
\newblock Micro-architecture independent analytical processor performance and
  power modeling.
\newblock In \emph{2015 IEEE International Symposium on Performance Analysis of
  Systems and Software (ISPASS)}, pages 32--41. IEEE, 2015.

\bibitem[Vasudevan et~al.(2021)Vasudevan, Jiang, Bieber, Singh, SHOJAEI, Ho,
  and Sutton]{vasudevan:2021:design2vec}
Shobha Vasudevan, Wenjie Jiang, David Bieber, Rishabh Singh, HAMID SHOJAEI,
  C.~Richard Ho, and Charles Sutton.
\newblock Learning semantic representations to verify hardware designs.
\newblock In A.~Beygelzimer, Y.~Dauphin, P.~Liang, and J.~Wortman Vaughan,
  editors, \emph{Advances in Neural Information Processing Systems}, 2021.

\bibitem[Wang et~al.(2019)Wang, Lee, Wei, and Brooks]{wang:2019:predictcpuspec}
Yu~Wang, Victor Lee, Gu-Yeon Wei, and David Brooks.
\newblock Predicting new workload or cpu performance by analyzing public
  datasets.
\newblock \emph{ACM Transactions on Architecture and Code Optimization (TACO)},
  15\penalty0 (4):\penalty0 1--21, 2019.

\end{thebibliography}
